\documentclass[12pt]{article}

\usepackage[english]{babel}

\usepackage[letterpaper,top=2cm,bottom=2cm,left=3cm,right=3cm,marginparwidth=1.75cm]{geometry}

\usepackage{amsmath}
\usepackage{graphicx}
\usepackage{subfigure}
\usepackage[T1]{fontenc}
\usepackage[utf8]{inputenc}
\usepackage{authblk}
\usepackage{float}
\usepackage{makecell}
\usepackage[colorlinks=true, allcolors=blue]{hyperref}
\providecommand{\keywords}[1]
{
  \small	
  \textbf{\textit{Keywords---}} #1
}

\title{YOLO-FaceV2: A Scale and Occlusion Aware Face Detector}
\author[1]{Ziping Yu}
\author[2]{Hongbo Huang\thanks{Corresponding Author: hhb@bistu.edu.cn}}
\author[3]{Weijun Chen}
\author[4]{Yongxin Su}
\author[5]{Yahui Liu}
\author[2]{Xiuying Wang}
\affil[1]{School of Instrument Science and Opto-electronic Engineering, Beijing Information Science and Technology University, Beijing, China}
\affil[2]{Computer School, Beijing Information Science and Technology University, Beijing, China}
\affil[3]{Data Algorithm, NIO, Shanghai, China}
\affil[4]{School of Mechanical and Electrical Engineering, Beijing Information Science and Technology University, Beijing, China}
\affil[5]{School of Information Management, Beijing Information Science and Technology University, Beijing, China}
\date{} 
\begin{document}
\maketitle

\begin{abstract}
\par
In recent years, face detection algorithms based on deep learning have made great progress. These algorithms can be generally divided into two categories, i.e. two-stage detector like Faster R-CNN and one-stage detector like YOLO. Because of the better balance between accuracy and speed, one-stage detectors have been widely used in many applications. In this paper, we propose a real-time face detector based on the one-stage detector YOLOv5, named YOLO-FaceV2.  We design a Receptive Field Enhancement module called RFE to enhance receptive field of small face,  and use NWD Loss to make up for the sensitivity of IoU to the location deviation of tiny objects. For face occlusion, we present an attention module named SEAM and introduce Repulsion Loss to solve it. Moreover, we use a weight function Slide to solve the imbalance between easy and hard samples and use the information of the effective receptive field to design the anchor. The experimental results on WiderFace dataset show that our face detector outperforms YOLO and its variants can be find in all easy, medium and hard subsets. Source code in \href{https://github.com/Krasjet-Yu/YOLO-FaceV2}{https://github.com/Krasjet-Yu/YOLO-FaceV2}
\end{abstract}

\section{Introduction}
\par
Face detection is an essential step in many face-related applications, such as face recognition, face verification and face attribute analysis, etc. With the booming of deep convolutional neural networks in recent years, the performance of face detectors has been greatly improved. Many high-performance face detection algorithms based on deep learning have been proposed. Generally, these algorithms can be divided into two branches.  One branch of typical deep-learning-based face detection algorithms \cite{1, 2, 3} uses cascading means of neural networks as feature extractors and classifiers to detect faces from coarse to fine. Despite their great success, it is important to note that cascade detectors suffer some drawbacks such as having difficulties in training and slow detection speed. The other branch is improved from general purpose object detection algorithms \cite{4, 5, 6}. General purpose object detectors take into account more common features and broader characteristics of objects. Therefore, task-specific detectors can share these information and then enforce the spectacular properties by special designs.  Some popular face detectors including YOLO \cite{7, 8, 9, 10}, Faster R-CNN \cite{5} and RetinaNet \cite{6} fall into this category. In this paper, inspired by YOLOv5 \cite{11}, TridentNet \cite{12} and Attention Network in FAN \cite{13}, we propose a novel face detector that achieves the state-of-the-art in one-stage face detection.
\par
Although deep convolutional networks have improved face detection remarkably, detecting faces with high variance in scale, pose, occlusion, expression, appearance, and illumination in realistic scenes remains great challenge. In our previous work, we proposed the YOLO-Face \cite{14}, an improved face detector based on YOLOv3 \cite{9}, which mainly focused on the problem of scale variance, design anchor ratios suitable for human face and utilized a more accurate regression loss function. The mAP of Easy, Medium, and Hard on the WiderFace \cite{15} validation set reached 0.899, 0.872, and 0.693, respectively. Since then variety of new detectors have been presented and the face detection performance has been significantly improved. However, for small objects, the one-stage detectors have to divide the search space with a finer granularity, so it is apt to cause the problem of imbalance of positive and negative samples \cite{16}. Furthermore, face occlusions \cite{13} in complex scenes affects the accuracy of the face detector remarkably. Aimed to address the problems of varying face scales, easy and hard sample imbalance and face occlusion, we propose a YOLOv5-based face detection method called YOLO-FaceV2.
\par
By carefully analyzing the difficulties encountered by face detectors and the shortcomings of YOLOv5 detector, we carry out the following solutions.
\par
\textbf{Multi scale fusion:} In many scenarios, there are usually different scale faces existing in the images, which is really difficult for them all to be detected by the face detector. Therefore, solving different scale faces is a very important task for face algorithms. Currently, the main method to solve the problem of varying scales is constructing a pyramid to fuse the multi-scale features of faces \cite{17, 18, 19, 20}. For example, in YOLOv5, FPN \cite{20} fuses the features of P3, P4 and P5 layers. However, for small-scale objects, the information can be easily lost after multi-layer convolutions, and the pixel information retained is very little, even in the shallower P3 layer. Therefore, increasing the resolution of the feature map can undoubtedly benefit the detection of small objects. 
\par
\textbf{Attention mechanism:} In many complex scenes, face occlusion often occurs, which is one of the main reasons for the accuracy decline of face detectors. To address this problem, some researchers try to use attention mechanism to facial feature extraction. FAN \cite{13} proposes a anchor-level attention. They suggest that the solution is to maintain the response value of the unobstructed region and to compensate the reduced response value of the obscured region through the attention mechanism. However, it doesn’t fully utilize the information between channels.
\par
\textbf{Hard Samples:} In one-stage detectors, many bounding boxes are not been filtered out iterately. So the number of easy samples in one-stage detectors is very large. During training, their cumulative contribution dominates the update of the model, leading to the overfit of the model \cite{16}. This is known as the problem of imbalanced samples. To deal with this problem, Lin et al. proposes Focal Loss to dynamically assign more weights to difficult sample examples \cite{6}. Similar to focal loss, Gradient Harmonizing Mechanism (GHM) \cite{21} suppresses the gradients from positive and negative simple samples to focus more on difficult samples. Prime Sample Attention (PISA) \cite{22} proposed by Cao et al. assigns weights to positive and negative samples according to different criteria. However, current hard sample mining methods have too many hyperparameters to be set, which is very inconvenient in practice.
\par
\textbf{Anchor design:} As pointed out in \cite{23}  a region in a CNN feature map has two types of receptive fields, the theoretical receptive field and the actual receptive field. It is experimentally shown that not all pixels in the receptive field respond equally, but obey a Gaussian distribution. This makes the anchor size based on the theoretical receptive field larger than its actual size, which makes it more difficult for the regression of bounding boxes. Zhang et. al designs the size of the anchors based on the effective receptive field in $S^3FD$ \cite{24}. And FaceBoxes \cite{25} designs the multiscale anchor to enrich the receptive fields and discretize anchors over different layers to handle faces of various scales. Therefore, the design of scales and ratios of the anchor boxes is very important which may greatly benefits the accuracy and convergence procedure of the model. 
\par
\textbf{Regression Loss:} Regression loss is used to measure the difference between the predicted bounding box and the ground truth bounding box. The commonly used regression loss functions in object detectors are L1/L2 loss, smooth L1 loss, IoU loss and its variants \cite{26, 27, 28, 29}. YOLOv5 takes IoU loss as its objective regression function. However, the sensitivity of IoU varies greatly for objects of different scales. It is readily comprehensible that, for small targets, a slight position deviation leads to a significant IoU decrease. Wang et al. \cite{30} proposes a small target evaluation method based on Wasserstein distance to effectively mitigate the effect of small target. However, their method performs not so significant for large targets.
\par
In this paper, to address the aforementioned problems, we design a new face detector based on YOLOv5. Our aim is to find an optimal combinatorial detector that effectively solves the problems of small faces, large scale variations, occluded scenes and imbalanced hard and easy samples. First, we fuse P2 layer information of FPN to obtain more pixel-level information and compensate the information of small face. However, in this way, the detection accuracy of large and medium targets will be slightly reduced because the output feature map perceptual field becomes smaller. To ameliorate this situation, we design Receptive Field Enhancement (RFE) for the P5 layer, which increases the receptive field by using dilated convolution. Second, inspired by FAN and ConvMixer \cite{31}, we redesign a multi-head attention network to compensate for the loss of occluded face response values. In addition, we also introduce Repulsion Loss \cite{32} to improve the recall of intra-class occlusions. Third, to mine hard samples, inspired by ATSS \cite{33}, we design the Slide weight function with adaptive thresholding to make the model focus more on hard samples during training. Fourth, in order to make the anchor more suitable for regression, we redesign the anchor size and proportion according to the effective receptive field and the proportion of the face. Fifth, we borrowed the Normalized Wasserstein Distance metric \cite{30} and introduced it into the regression loss function to balance the shortage of IoU in predicting small faces.
\par
In summary, we propose a new face detector YOLO-FaceV2, in which the highlighted contributions are as follows.
\par
1. For detecting multiscale faces, the perceptive field and resolution are key factors. Therefore, we design a receptive field enhancement module (called RFE) to learn different receptive fields of the feature map and enhance the feature pyramid representation.
\par
2. We classify the face occlusions into two categories, i.e., the occlusion between different faces, and the occlusion of faces by other objects. The former makes the detection accuracy very sensitive to NMS thresholds which leads to missed detection. We use Repulsion Loss to face detection which penalizes the predicted box for shifting to the other ground-truth objects and requires each predicted box to keep away from the other predicted boxes with different designated targets to make the detection results less sensitive to NMS. The latter causes feature disappearance leading to inaccurate localization, and we design the attention module SEAM to enhance the learning of face features.
\par
3. To address the problem of imbalance between hard and easy samples, we weight the easy and hard samples according to the IoU. To reduce hyperparameter tuning, we set the mean value of IoU of all candidate positive samples with ground-truth as the dividing line between positive and negative samples. And we design a weighted function named Slide to give higher weight to hard samples which is helpful for the model to learn more difficult features. The details of this function will be presented in sections 3-5.
\par
The rest of the paper is arranged as follows: in Section 2 we review the related literature in this area; in Section 3 we describe the model structure in detail, and the main improvisions including the receptive field enhancement module, the attention module, the adaptive sample weighting function, the anchor design, the Replusion Loss and the Normalized Gaussian Wasserstein Distance (NWD) Loss, respectively; in Section 4 we describe the experiments and the according analysis of the results, including ablation experiments and comparisons with other models; and in Section 5 we summarize our work and give some advice about future research.

\section{Related Works}
\textbf{Face Detection.} Face detection has been a hot research area in computer vision for decades. In the early years of deep learning, face detection algorithms usually use neural networks to automatically extract image features  for classification. CascadeCNN \cite{1} proposes a cascaded structure with three stages of carefully designed deep convolutional networks that predicts face and landmark location in a coarse-to-fine manner.  MTCNN \cite{2} develops a similar cascade architecture to jointly align the face landmarks and detect the face locations. PCN \cite{3} uses an angle prediction network to correct faces and improve the face detection accuracy. But early deep-learning-based face detection algorithms have some drawbacks such as tedious training, local optimum, slow detection speed, and low detection accuracy, etc.
\\
Current face detection algorithms are mainly improved by inheriting the advantages of generic object detection algorithms, such as SSD \cite{4}, Faster R-CNN \cite{5}, RetinaNet \cite{6}, etc. CMS-RCNN \cite{34} uses Faster R-CNN as backbone and introduces contextual information and multi-scale features to detect faces. Zhang et al. \cite{25} designs a lightweight network based on SSD structure, named FaceBoxes, to quickly shrink the feature size by 32x down-sampling, and uses a multi-scale network module to enhance the features in both network width and depth dimensions. SRN \cite{35}, which is improved on the generic object detection algorithm RefineDet \cite{36} and RetinaNet \cite{6}, achieves high performance by introducing two-stage classification and regression, and designs a multi-branch module to enhance the effect of receptive fields.
\\
\textbf{Scale-invariance.} As one of the most challenging problems in face detection, large face scale variations in complex scenes has an important impact on the accuracy of the detector. The multi-scale detection capability mainly depends on the scale-invariance features, and many works address this problem to extract features more accurately and effectively \cite{13, 24, 37, 38}. For small objects detection, using fewer down-sampling layers and dilated convolution can significantly improve the detection performance \cite{39, 40}. Another way to bridge this problem is using more anchors. Anchor can provide good priori information, thus using denser anchors and corresponding matching strategies can effectively improve the quality of object proposals \cite{24, 25, 37, 40}. Multi-scale training can be helpful to construct the image pyramids and increase the sample diversity, which is a simple but effective method to improve the performance of multi-scale object detection. On the other hand, the receptive fields will increase and the semantic information get richer accordingly, however, the spatial information may be missing correspondingly. A natural idea is to fuse deep semantic information with shallow features, such as \cite{20, 41, 42}. Besides, SNIP \cite{43} and TridentNet \cite{12} also provide new ideas to solve the multi-scale problem, which will be discussed in detail in the following sections.
\\
\textbf{Occlusion problem.} Crowding faces and the following occlusion problem give rise to partial data and lack of information about the occluded faces, because some regions are invisible or the boundaries are blurred, which can easily cause missed detection and low recalls. Some works have demonstrated that contextual information is helpful for face detection to alleviate the occlusion problem. SSH \cite{37} uses means of simple convolution layers to incorporates context by enlarging the window around the candidate proposals. FAN \cite{13} proposes an anchor-level attention to detect the occluded faces by highlighting the features from the face region. PyramidBox \cite{44} designs a context-sensitive predict module in which they replace the convolution layers of context module in SSH by the residual prediction module of DSSD. RetinaFace \cite{45} applies independent context modules on five feature pyramid levels to increase the receptive field and enhance the rigid context modelling power. The above methods have achieved good results in the occlusion problem. Therefore, using context information to improve the effectiveness of the occluded regions is a feasible direction which is worth further exploration.
\\
\textbf{Imbalance of easy and hard samples.} For one-stage face detection, the number of easy samples is very large, and they dominate the variation of losses so that the model can only learn the features of easy samples and ignores the learning of hard samples. To address this problem, the OHEM \cite{46} algorithm selects the difficult samples according to the sample loss and applies the loss of difficult samples to the training in stochastic gradient descent. In response to the problem of ignoring easy samples in the OHEM algorithm, Focal Loss \cite{6} makes better use of all samples by weighting them and obtains higher accuracy. This idea is also followed by SRN \cite{35}. Faceboxes \cite{25} sorts the samples according to their IoU loss, and controls the ratio of positive to negative samples to be less than 1:3. Although the above methods can effectively solve the problem of sample imbalance, they also artificially introduce some hyperparameters, which increase the difficulty of adjusting. Therefore, we design a sample balance function with adaptive parameters.

\section{YOLO-FaceV2}
\subsection{Network Architecture}

YOLOv5 is an excellent general object detector. We introduce YOLOv5 into the face detection field and try to solve the problems of small faces and face occlusion, etc.
\par
\begin{figure}[ht]
\centering
\includegraphics[width=0.9\textwidth]{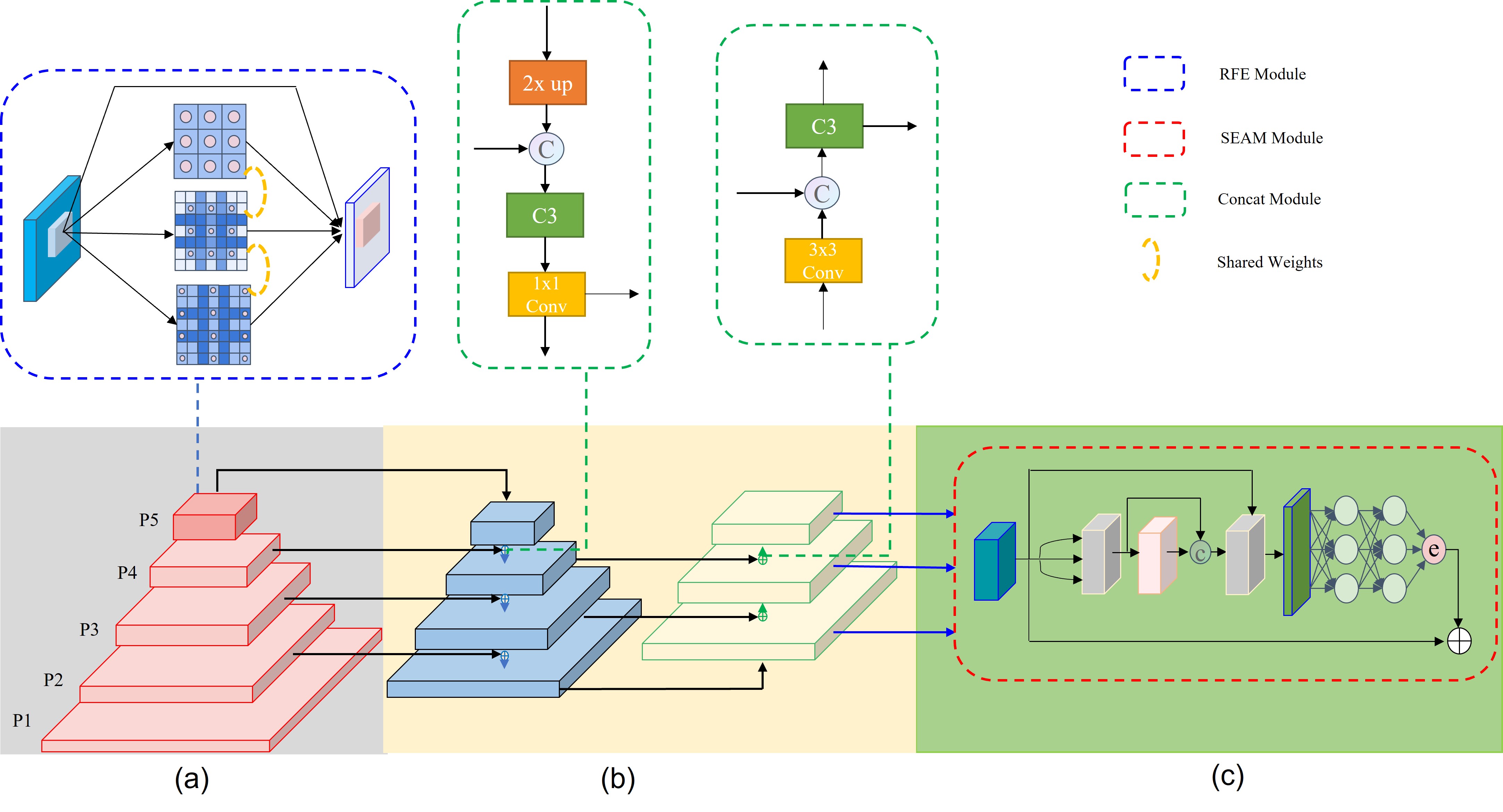}
\caption{\label{Fig.1}Network architecture of YOLO-FaceV2. (a) Backbone: a feed-forward CSPDarknet53 architecture extracts the multi-scale feature maps. To expand the receptive field, the CSP block in P5 is replaced by the receptive field enhancement module (RFE) which is shown in the blue dotted box.  (b) Neck: a Spatial Pyramid Pooling (SPP) separates out the most significant context features and increases the receptive field and a Path Aggregation Network (PAN) aggregates parameter from different backbone levels for different detector levels. To compensate for increased receptive field loss of resolution, the P2 layer is fused into PAN which is shown between (a) and (b). (c) a Separated and Enhancement Attention Module (SEAM) uses the relationship between feature maps to recall occluded features which is shown in the red dotted box.}
\end{figure}

The architecture of our YOLO-FaceV2 detector is shown in Figure \ref{Fig.1}. It consists of three parts: the backbone structure, the neck and the heads. We take CSPDarknet53 as our backbone and replace the Bottleneck with RFE module in P5 layer to fuse multi-scale features. In the neck part, we maintain the structure of SPP \cite{47} and PAN \cite{48}. In addition, in order to improve the ability of target position perception, we also integrate the P2 layer into the PAN. The heads are used to classify the category and regress the location of the target. We also add a special branch into the heads to enhances the model’s ability of occlusion detection. 
\par
In Figure \ref{Fig.1} (a), the red part on the left is the backbone of the detector, which is composed of CSP blocks and CBS blocks. It is mainly used to extract the features of the input images. And the RFE module is added to expand the effective receptive field and enhance the fusion capability of multi-scale in the P5 layer. In Figure \ref{Fig.1} (b), the blue and yellow parts on the right are called neck layers, which consists of SPP and PAN. We additionally fuse the features of the P2 layer to improve the ability of more accurate target localization. In Figure \ref{Fig.1} (c), we introduce the separated and enhancement attention module(SEAM) to strengthen the responsiveness of occluded faces after the output part of the neck layer.

\subsection{Scale-Aware RFE Model}
\begin{figure}[ht]
\centering
\includegraphics[width=0.9\textwidth]{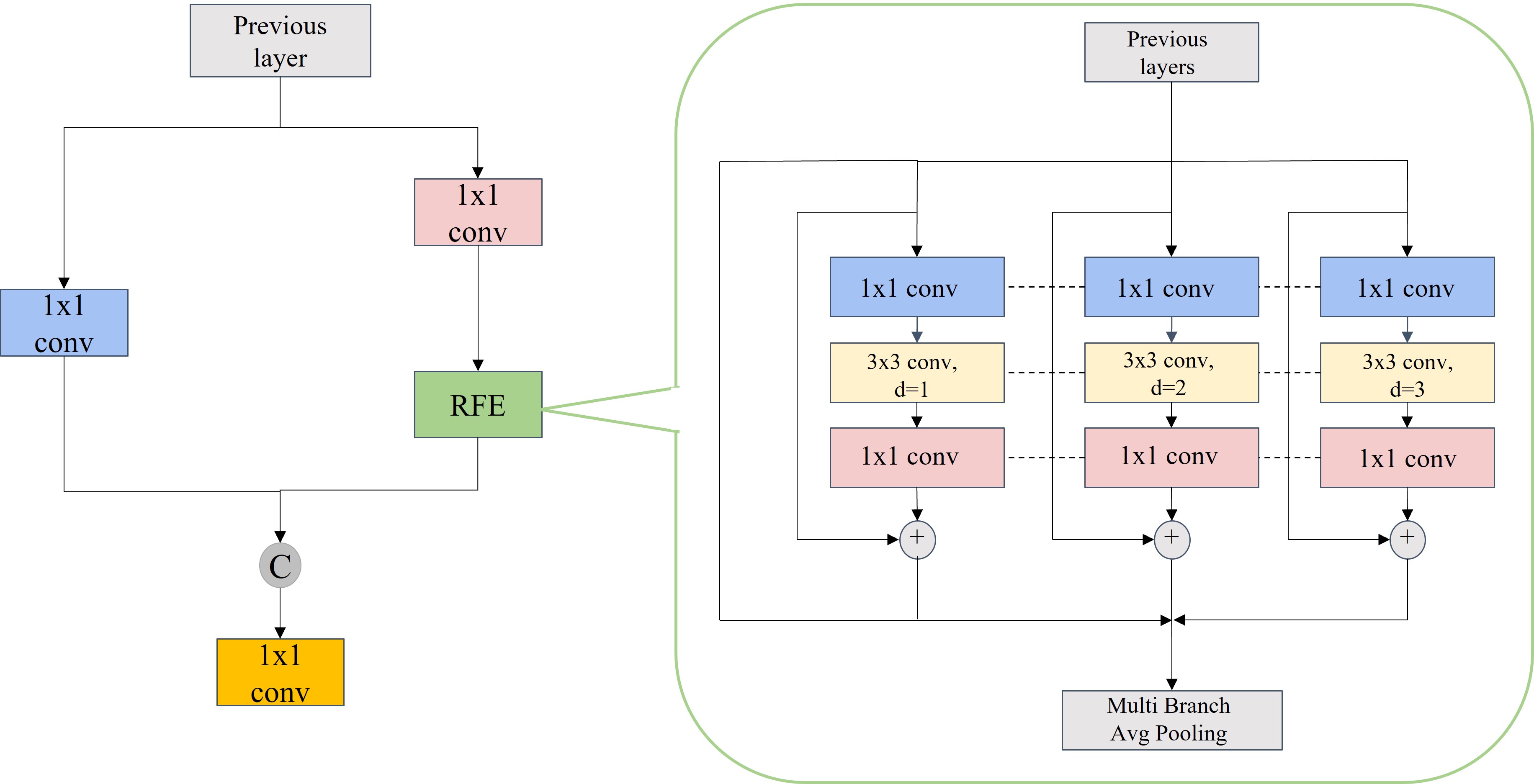}
\caption{\label{Fig.2}Modified CSP block and RFE module. For CSP block in P5, we replace bottleneck with RFE. The right figure shows the detailed architecture of the RFE. It consists of 1x1 convolution, 3x3 convolution with different dilated rate and average pooling layer.}
\end{figure}

Since different size of receptive fields means different capability of capturing long-range dependency, we design the RFE module to sufficiently make use of the advantage of receptive fields in the feature map by using dilated convolution. Inspired by TridentNet, we use four branches of different rates of dilated convolution to capture multi-scale information and different ranges of dependency. All the branches have sharing weights, the only difference is their distinctive receptive fields. On the one hand, it reduces the amount of parameters and thus the risk of potential overfitting. On the other hand, it can make full use of each sample. The proposed RFE module can be divided into two parts: multi branches based on dilated convolutions and the gathering \& weighting layer, as shown in Figure \ref{Fig.2}. The multi-branch part takes 1,2 and 3 as the rates of different dilated convolutions separately, which all use fixed convolutional kernel size 3x3. Furthermore, we add a residual connection to prevent the problem of gradient explosion and disappearance during training. The gathering and weighting layer is used to gather information from different branches and weight every branch of the features.  The weighting operation is used to balance the representation of different branches. 
\par
To make it clear, we replace the bottleneck of C3 module in YOLOv5 with RFE module to increase the receptive field of feature map, so as to improve the accuracy of multi-scale target detection and recognition, as shown in Figure \ref{Fig.2}.

\subsection{Occlusion-Aware Repulsion Loss}
Intra class occlusion may cause face A contains the features of face B, resulting in a higher false detection rate. The introduction of the repulsion loss can effectively alleviate this problem through repulsion. The repulsion loss is divided into two parts: RepGT and RepBox. The function of RepGT Loss is to make the current bounding box as far away from the surrounding ground truth box as possible. The surrounding ground truth box here refers to the face label with the largest IoU with the face except for the object to be returned by the bounding box itself. The formula of RepGT loss function is as follows:
\begin{equation}
L_{\text {RepGT }}=\frac{\sum_{P \in \mathcal{P}_{+}} \operatorname{Smooth}_{l n}\left(\operatorname{Io} G\left(P, G_{R e p}^{P}\right)\right)}{\left|\mathcal{P}_{+}\right|} \label{1}
\end{equation}\\
where
\begin{equation}
\operatorname{Smooth}_{l n}=\left\{\begin{array}{ll}
-\ln (1-x) & x \leq \sigma \\
\frac{x-\sigma}{1-\sigma}-\ln (1-\sigma) & x>\sigma
\end{array}\right.
\label{2}
\end{equation}
\par
P in the formula is the face prediction frame, $G_{\operatorname{Re} p}^{P}$ is the ground truth with the largest IoU around the face. The overlap between P and $G_{\operatorname{Re} p}^{P}$ is defined as intersection over ground truth (IoG): $\operatorname{IoG}(P, G) = \frac{\operatorname{area}(P \cap G)}{\operatorname{area}(G)}$ and $\operatorname{IoG}(B, G) \in[0,1]$. $\operatorname{Smooth}_{l n}$ is (0, 1)  a continuously differentiable. ln function, $\sigma \in[0,1)$ is a smoothing parameter to adjust the sensitivity of repulsion loss to outliers.
\par
The purpose of RepBox loss is to make the prediction frame as far away from the surrounding prediction frame as possible and reduce the IOU between them, so as to avoid one of the prediction frames belonging to two faces being suppressed by NMS. We divide the prediction frame into multiple groups. Assuming there are g individual faces, the division form is shown in Eqn \ref{3}. The prediction frames between the same groups return to the same face label, and the prediction frames between different groups correspond to different face labels.
\begin{equation}
\rho_{+}=\rho_{1} \cap \rho_{2} \cap \ldots \cap \rho_{|g|}
\label{3}
\end{equation}
\par
Then, for the prediction box between different groups $p_{i}$ and $p_{j}$, we hope that we get the smaller the overlap area between $p_{i}$ and $p_{j}$. RepBox also uses ${Smooth}_{ln}$ as an optimization function. The overall loss function is as follows:
\begin{equation}
L_{R e p B o x}=\frac{\sum_{i\neq j} S m o o t h_{\ln }\left(\operatorname{IoU}\left(B^{p_{i}}, B^{p_{j}}\right)\right)}{\sum_{i \neq j} 1\left[\operatorname{IoU}\left(B^{p_{i}}, B^{p_{j}}\right)>0\right]+\epsilon}
\label{4}
\end{equation}

\subsection{Occlusion-Aware Attention Network}
\begin{figure}[ht]
\centering
\includegraphics[width=0.8\textwidth]{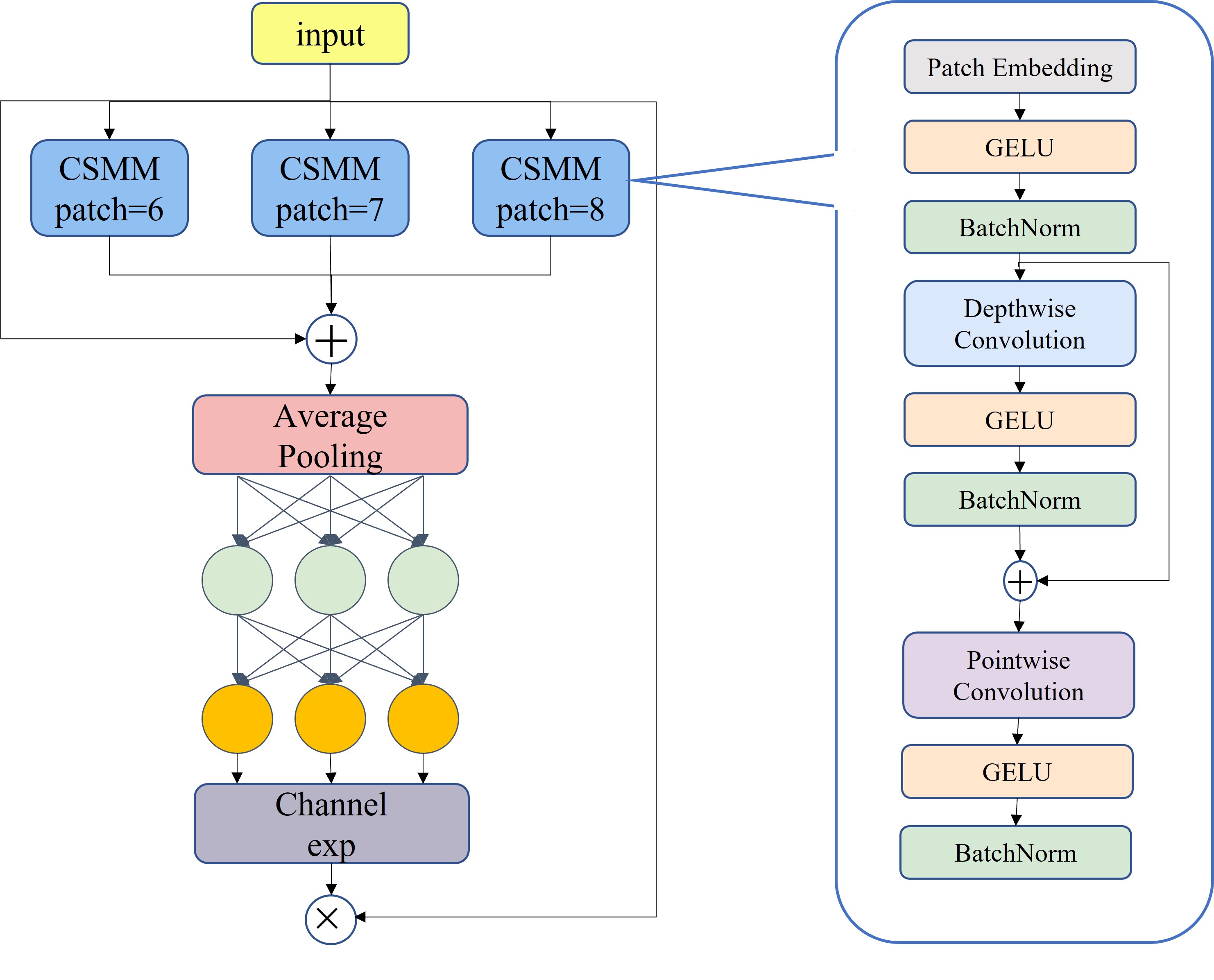}
\caption{\label{Fig.3}Illustration of SEAM. The left is the architecture of the SEAM, the right part is the structure of CSMM(channel and spatial mixing module).  The CSMM utilizes different patch to multi-scale feature and uses depth separable convolution to learn the correlation of spatial dimensions and channels.}
\end{figure}

Inter class occlusion will cause alignment error, local aliasing and feature missing. We add the multi head attention network, namely the SEAM module (see Figure \ref{Fig.3}), in which we have three purposes:  realize multi-scale face detection, emphasize the face area in the image and weaken the background area oppositely. The first part of SEAM is the depth separable convolution with residual connection. Depth separable convolution is operated depth by depth, that is, the convolution separated channel by channel. Although depth separable convolution can learn the importance of different channels and reduce the amount of parameters, it ignores the information relationship between channels. To make up for this loss, the outputs of different depth convolutions are subsequently combined by point-by-point (1x1) convolutions. Then a two-layer full connection network is used to fuse the information of each channel, so that the network can strengthen the connection between all channels. It is hoped that this model can make up for the aforementioned loss under occlusion scenarios through the relationship between the occluded face and the unobstructed face learned in the previous step. The output logits learned by the full connection layer is then processed by an exponential function to expand the value range from [0, 1] to [1, e].  This exponential normalization provides a monotonic mapping relationship that makes the results more tolerant to position error. Finally, the output of the SEAM module is used as attention that is multiplied by the original features, so that the model can deal with the face occlusion more effectively.

\subsection{Sample weighting function}
\begin{figure}[ht]
\centering
\includegraphics[width=0.5\textwidth]{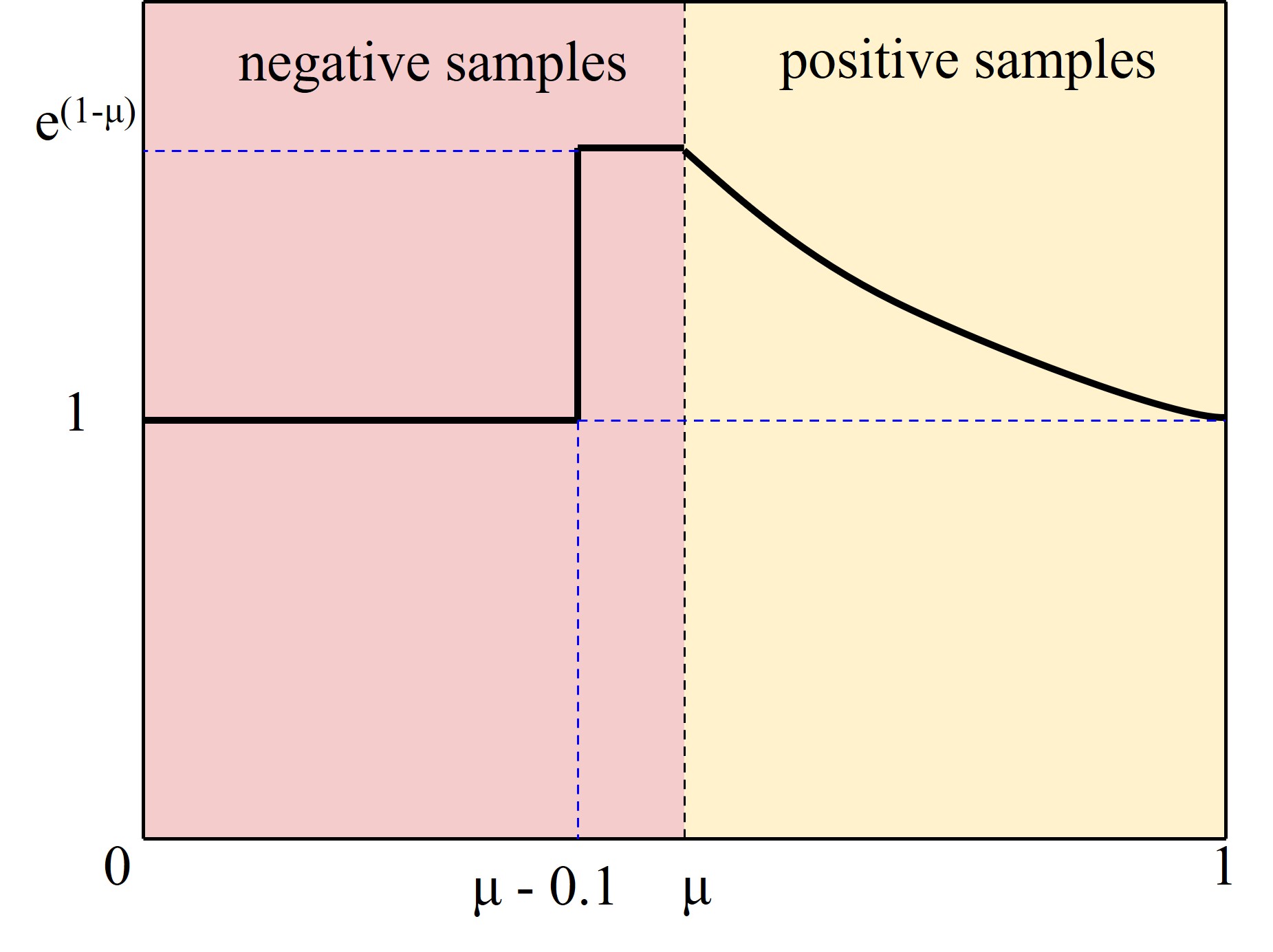}
\caption{\label{Fig.4}We propose a novel loss we term the Slide Loss that adaptively learns the positive and negative sample threshold parameters $\mu$. Setting high weights near $\mu $ increases the relative loss for hard-classified examples, putting more focus on hard, misclassified examples.}
\end{figure}

The sample imbalance problem, i.e., in most cases the number of easy samples are quite large while hard samples are relatively sparse, has attracted a lot of attention. In our work, we design a Slide Loss function which looks like a "slide" to address this problem. The distinction between easy and hard samples is based on the IoU size of the prediction box and groundtruth box. To reduce hyperparameters, we take the average of the IoU value of all bounding boxes as threshold value $\mu$, those less than $\mu$ are taken as negative samples, and positive samples for the values greater than $\mu$. However, samples near the boundaries often suffer from large losses due to unclear classifications. We hope that the model can learn to optimize these samples and use these samples more sufficiently to train the network. Nevertheless, the number of such samples is relatively small. Therefore, we try to assign higher weights to the difficult samples. We first divide the samples into positive and negative samples by the parameter $\mu$. Then, we emphasis on the samples at the boundary by a weighting function Slide, as shown in Figure \ref{Fig.4}. The Slide weighting function can be expressed as Eqn \ref{5}.
\begin{equation}
f(x)=\left\{\begin{array}{ll}
1 & x \leq \mu-0.1 \\
e^{1-\mu} & \mu<x<\mu-0.1 \\
e^{1-x} & x \geq u
\end{array}\right.
\label{5}
\end{equation}

\subsection{Anchor Design Strategy}
\begin{figure}[ht]
\centering
\includegraphics[width=0.5\textwidth]{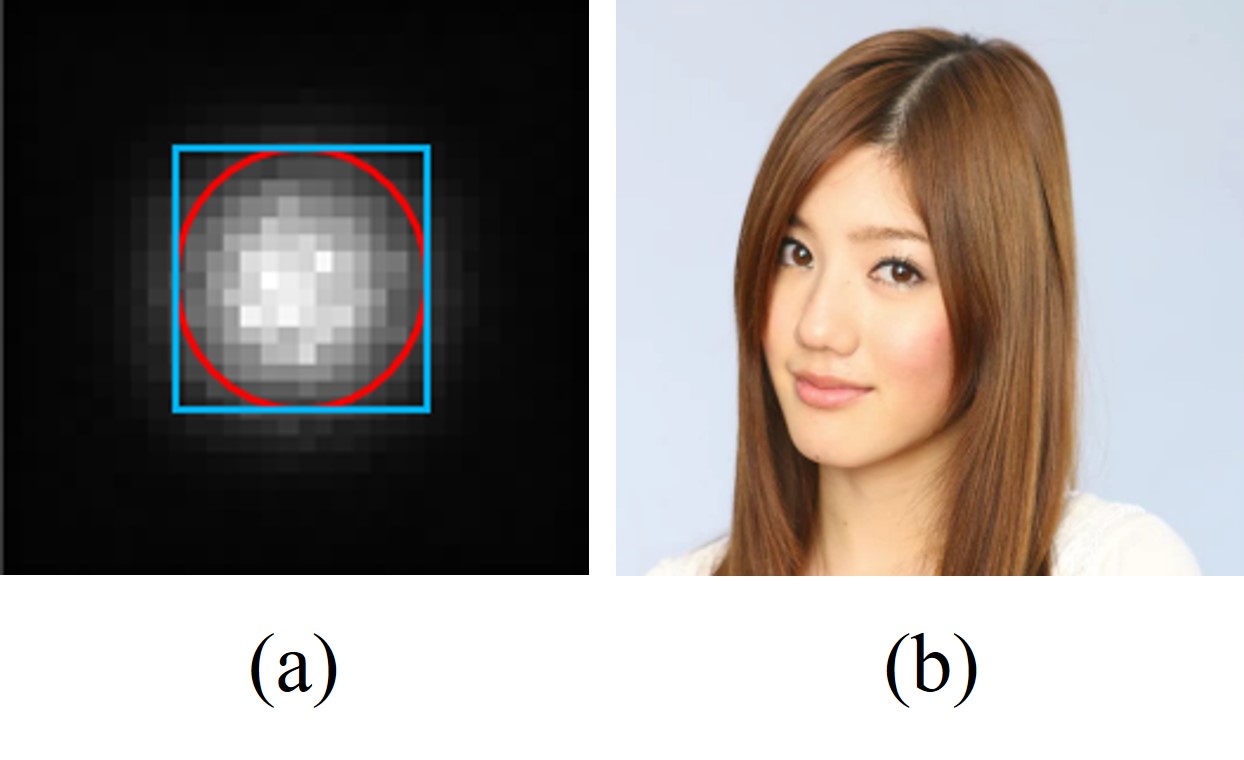}
\caption{\label{Fig.5}(a) \textbf{Effective receptive field:} The whole black box is the theoretical receptive field (TRF) and the while circle with Gaussian distribution is the effective receptive field (ERF). The figure is from \cite{23}. (b) \textbf{A special example:} The whole box is the origin anchor box setting by TRF and the blue box is the new anchor estimated by the red circle which is the ERF.}
\end{figure}

\begin{table}[ht]
\centering
\caption{\label{Tab.1}The three detection layers and nine anchor scales.}
\begin{tabular}{c|c|c|c}\\
Layer & Stride & Ratio & Anchor \\\hline
P2 & 4 & 1.2 & [16, 20.16, 25.40] \\
P3 & 8 & 1.2 & [32, 40.32, 50.80] \\
P4 & 16 & 1.2 & [64, 80.63, 101.59]
\end{tabular}

\end{table}
Anchor design strategy is critical in face detection. In our model, each of the three detection heads is associated with a specific anchor scale. The design of anchors includes the ratio of width to height and the size of anchors which are designed according to the stride of P2, P3 and P4(see Table~\ref{Tab.1}). For the ratio of width to height, we calculate the statistics from the WiderFace train set based on the ground-truth face ratio. Here in face detection, we set the aspect ratio to 1:1.2 according to the statistics. For the size of anchors, we design it according to the receptive field of each layer, which can be calculated by the number of convolution and pooling layers. However, not every pixels in the theoretical receptive field contributes the same to the final output. In general, the central pixel has a greater influence than the peripheral pixel, as shown in Figure \ref{Fig.5} (a). In other words, only a small part of the area has an effective influence on the output value. The actual effect can be equivalent to an effective receptive field. According to this hypothesis, in order to match the effective receptive field, the anchor should be significantly smaller than the theoretical receptive field (see the specific example in Figure \ref{Fig.5} (b)). Therefore, we redesigned the initial anchor size as shown in the Table~\ref{Tab.1}.

\subsection{Normalized Gaussian Wasserstein Distance}
Normalized Wasserstein Distance named NWD is a new evaluation method for small target detection. Firstly, the bounding box is modeled as a two-dimensional Gaussian distribution, and the similarity between the predicted target and the real target is calculated through their corresponding Gaussian distribution, that is, the normalized wasserstein distance between them is calculated according to Eqn \ref{6}. For the detected targets, whether they overlap or not, they can be measured by the distribution similarity. The NWD is not sensitive to the scale of the targets and thus is more suitable for measuring the similarity between small targets. In our regression loss function, we add a NWD Loss to make up for the disadvantage of the IoU loss for small target detection. But we still keep the IoU loss because it is suitable for large object detection.
\begin{equation}
N W D\left(\mathcal{N}_{a}, \mathcal{N}_{b}\right)=\exp \left(-\frac{\sqrt{W_{2}^{2}\left(\mathcal{N}_{a}, \mathcal{N}_{b}\right)}}{C}\right)
\label{6}
\end{equation}
\begin{equation}
W_{2}^{2}\left(\mathcal{N}_{a}, \mathcal{N}_{b}\right)=\left\|\left(\left[c x_{a}, c y_{a}, \frac{w_{a}}{2}, \frac{h_{a}}{2}\right]^{\mathrm{T}},\left[c x_{b}, c y_{b}, \frac{w_{b}}{2} \cdot \frac{h_{b}}{2}\right]^{\mathrm{T}}\right)\right\|_{2}^{2}\label{7}
\end{equation}
Where C is a constant closely related to the data set, $W_{2}^{2}\left(\mathcal{N}_{a}, \mathcal{N}_{b}\right)$ is a distance measure, and $N_{a}$ and $N_{b}$ are Gaussian distributions modeled by $A=\left(c x_{a}, c y_{a}, w_{a}, h_{a}\right)$  and $ B=\left(c x_{b}, c y_{b}, w_{b}, h_{b}\right)$.

\section{Experiments}
In this part, we conduct comprehensive ablations of our proposed method, including the effectiveness of our attention module, multi-scale fusion pyramid structure and loss function design. Then, we compare the performances between our proposed detector and other SOTA face detectors.
\subsection{Dataset}
We evaluated our model on the WiderFace dataset, which has 32203 images, including more than 400k faces. It consists of three parts: 40\% for training set, 10\% for verification set and 50\% for test set. The results of training set and verification set can be obtained from the official website of WiderFace. According to the difficulties, the dataset can be divided into three parts: easy, medium and hard. Among them, hard subset is the most challenging, and its performance can better reflect the effectiveness of face detector. We trained our model on the WiderFace training set and evaluated it on the validation set and test set.

\subsection{Training}
We use YOLOv5 as our baseline and the methods are implemented by PyTorch.The optimizer we used is SGD with momentum. The initial learning rate is set to 1e-2, the final learning rate is 1e-3, and the weight decay is set to 5e-3. A momentum 0.8 is used in first 3 warming-up epochs. After that, the momentum is 0.937.  The IoU for the NMS is set to 0.5. We train the model on 1080ti which have 4 CPU workers. The fine-tuning consumes 100 iterations with a batch size of 16 images.

\subsection{Ablation Study}
In this section, we conduct comprehensive experiments of each module on the WiderFace dataset to evaluation their affections on the performance of the model. Then the modules are combined and analyzed one-by-one. Moreover, all the loss functions are also evaluated.
\begin{table}[ht]
\scriptsize
\centering
\caption{\label{Tab.2}Ablation study results on the WiderFace validation dataset}
\begin{tabular}{c|c|c|c|c|c|c|c|c|c|c|c}\\
SEAM & PAN+P2 & RFE & Slide & Anchor & \makecell[c]{NWD \\ Loss} & RPLoss & Easy & Medium & Hard & \makecell[c]{Params \\ (M)} & \makecell[c]{Flops \\ (G)} \\\hline
&&&&&&&94.65&93.00&83.30&7.063&16.4 \\
$\surd$ &&&&&&&95.53&93.82&84.36&7.464&17.1 \\
& $\surd$ &&&&&&93.67&92.14&83.87&6.101&17.1 \\
& $\surd$ & $\surd$ &&&&&95.06&93.60&85.47&5.097&17.1 \\
&&&$\surd$&&&&95.13&93.41&83.67&-&17.1  \\
&&&&$\surd$&&&94.89&93.75&84.20&-&-  \\
&&&&&$\surd$&&94.62&92.87&83.31&-&-  \\
&&&&&&$\surd$&95.27&93.63&83.80&-&-  \\
$\surd$&$\surd$&$\surd$&&&&&95.06&93.64&85.57&5.201&17.9  \\
$\surd$&$\surd$&$\surd$&$\surd$&&&&95.34&93.85&85.66&&  \\
$\surd$&$\surd$&$\surd$&$\surd$&$\surd$&&&96.22&94.79&85.82&&  \\
$\surd$&$\surd$&$\surd$&$\surd$&$\surd$&$\surd$&&96.30&94.99&85.94&& \\
$\surd$&$\surd$&$\surd$&$\surd$&$\surd$&$\surd$&$\surd$&\textbf{98.78}&\textbf{97.39}&\textbf{87.75}&&18.2  \\
\end{tabular}
\end{table}

\subsubsection{SEAM Block}
Our proposed SEAM block is the attention network. By using this block, we make up for the response loss of the occluded face by strengthening the response of unobstructed faces. The results are shown in the second row of Table~\ref{Tab.2}. As we can see, the accuracy increases by 0.88, 0.82 and 1.06 on the easy, medium and hard subset validation sets respectively.

\subsubsection{Multi-scale feature fusion}
Firstly, we fuse P2 layer features on the basis of PAN, so that the fused feature map contains more information of small targets. According to the third row of Table~\ref{Tab.2}, it can be observed that the hard subset has increased by 0.57. In order to make up for the shortage of limited receptive field in the output feature map of the neck layer, which leads to the decline of detection accuracy of large and medium-sized targets, we applied the designed receptive field enhancement module, and used dilated convolutions, whose dilating rate are 1, 2 and 3, respectively, to improve the effect of long range dependency. The effect is shown in the fourth row of Table~\ref{Tab.2}. The accuracy has increased by 0.5, 0.6 and 2.17 respectively.

\subsubsection{Slide Loss}
The main purpose of the Slide loss function  is to make the model focus more on hard samples. According to the results in the fifth row of the table, the Slide function slightly improves the model of the model on the medium and hard subsets.

\subsubsection{Anchor Design}
The ratio and size of anchor are closely related to the effective receptive field. Different models have different effective receptive fields. According to the effective receptive field and face shape characteristics, the performance impact of designed anchors is shown in the sixth row of Table~\ref{Tab.2}. It improves by 0.24, 0.75, 0.9 on the easy, medium and hard datasets respectively. As we would expect, properly designed anchors can recall more small face targets.

\subsubsection{NWD Loss}
We have first adopted NWD instead of IOU as the regression loss. However, the result is not improved. Therefore, we choose to retain the IoU Loss and improve the robustness of our model to small target detection by adjusting the proportional relationship between them. Because the experimental results show that for large and medium-sized targets, the effect measured by IoU is better than NWD, and NWD can effectively improve the detection accuracy of small targets. The result show in Table~\ref{Tab.3}:
\begin{table}[ht]
\centering
\small
\caption{\label{Tab.3}Comparison of our YOLO-FaceV2 and existing face detector on the WiderFace validation dataset.}
\begin{tabular}{c|c|c|c|c|c}\\
IoU & NWD & Easy & Medium & Hard & Epochs \\\hline
1 & 0 &94.4&92.74&82.91&20 \\
0 & 1 &81.13&84.4&75.77&20 \\
0.5 & 0.5 &94.62&92.87&83.31&20 \\
0.4 & 0.6 &91.13&90.38&80.11&20 \\
0.6 & 0.4 &92.87&91.39&80.91&20
\end{tabular}

\end{table}

\subsubsection{Balance of RepGT and RepBox}
Inspired by the solution of occlusion in pedestrian detection, we add the Repulsion Loss to face detection, and analyze the different face occlusion thresholds to make this loss function applicable to face detection. According to the results in the eighth row of the table, the repulsion loss function improves the model accuracy by 0.71, 0.63 and 0.5 on easy, medium and hard subsets.

\subsection{Comparisons with Existing Face Detectors}

\begin{table}[ht]
\centering
\caption{\label{Tab.4}Comparison of our YOLO-FaceV2 and existing face detector on the WiderFace validation dataset.}
\begin{tabular}{c c c c c}\\
\hline
Method & Detector & Easy & Medium & Hard \\\hline
Faster R-CNN &&&& \\\hline
	&CMS-RCNN	&0.899	&0.874	&0.624\\
	&HR	        &0.925	&0.91	&0.806\\
	&Face R-CNN	&0.937	&0.921	&0.831\\
	&FDNet	    &0.959	&0.945	&0.879\\\hline
SSD          &&&&\\\hline
	&SFD	    &0.937	&0.925	&0.859\\
	&SSH	    &0.931	&0.921	&0.845\\
	&PyramidBox	&0.961	&0.95	&0.889\\
	&DSFD	    &0.966	&0.957	&0.904\\
	&SFDet	    &0.954	&0.945	&0.888\\\hline
RetinaNet    &&&&\\\hline
	&FAN	    &0.952	&0.94	&0.9\\
	&SRN	    &0.964	&0.952	&0.901\\
	&DFS	    &0.969	&0.959	&0.912\\
	&RetinaFace	&0.969	&0.961	&0.918\\
	&RefineFace	&0.971	&0.962	&\textbf{0.92}\\\hline
YOLO         &&&&\\\hline
	&YOLO-FaceV1	&0.899	&0.872	&0.693\\
	&YOLO5Face	    &0.963	&0.956	&0.913\\
	&YOLO-FaceV2	&\textbf{0.987}	&\textbf{0.972}	&0.877

\end{tabular}
\end{table}

\begin{figure}[H]
\centering
\subfigure[Easy]{
\includegraphics[width=11.5cm]{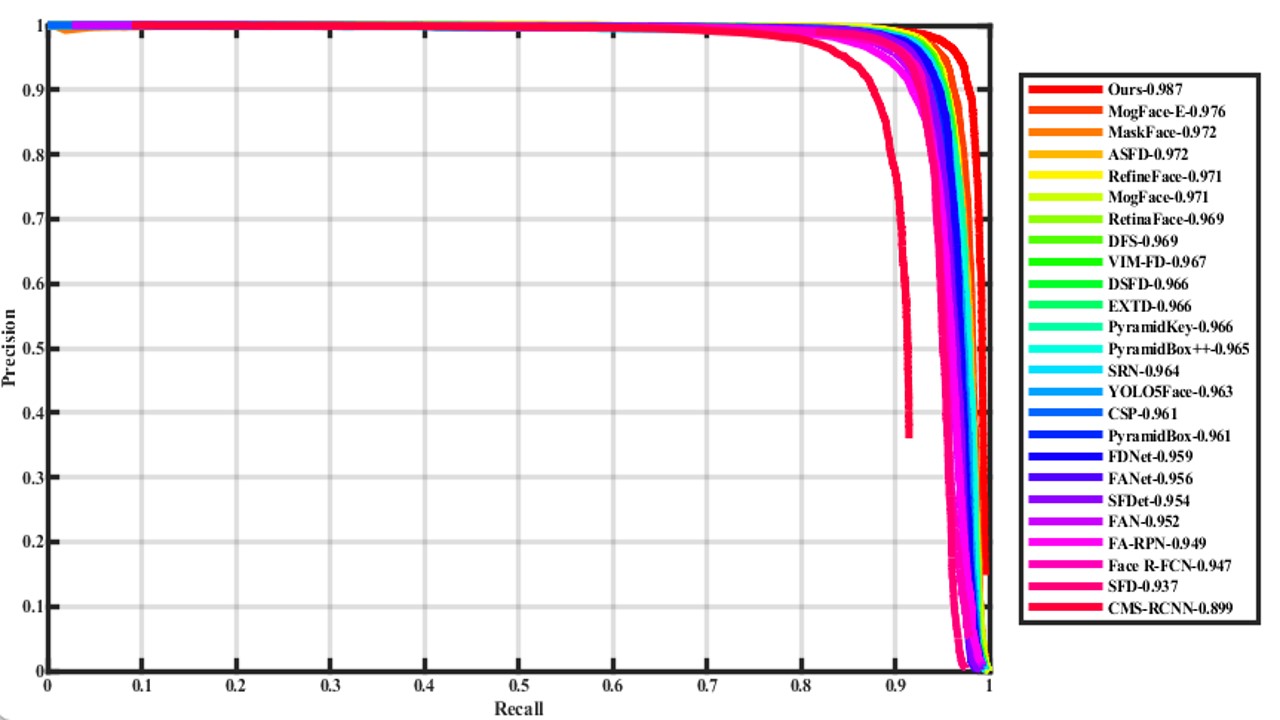}
}
\quad
\subfigure[Medium]{
\includegraphics[width=11.5cm]{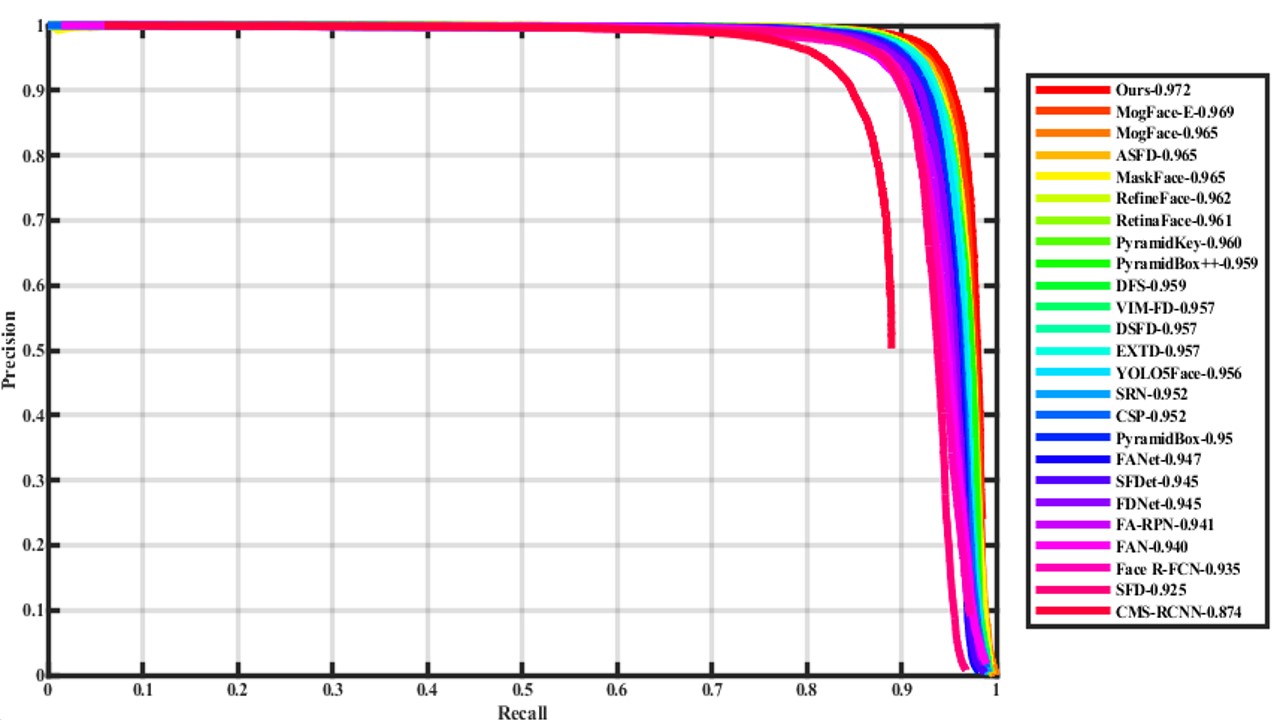}
}
\quad
\subfigure[Hard]{
\includegraphics[width=11.5cm]{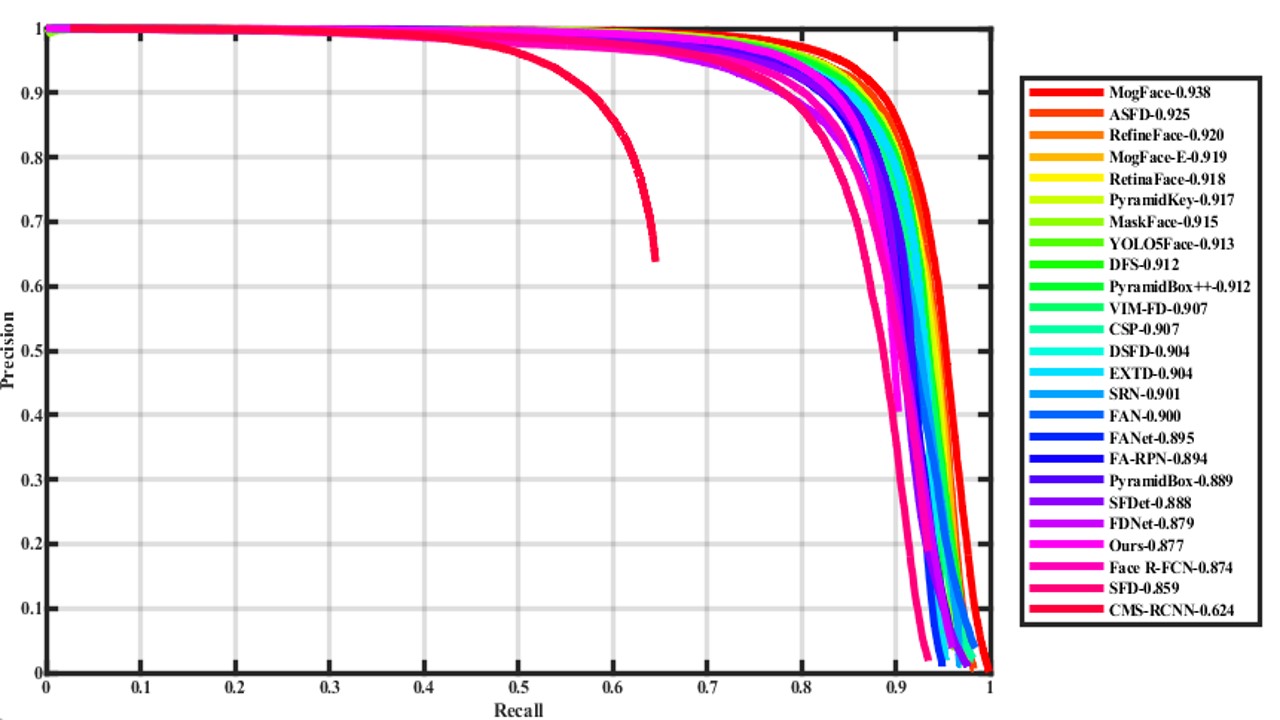}
}
\caption{\label{Fig.6}Detection results on the WiderFace validation dataset. (a): results on the ‘Easy’ dataset (b): results on the ‘Medium’ dataset (c): results on the ‘Hard’ dataset}
\end{figure}
We mainly compare with various excellent face detectors presented recently. Table~\ref{Tab.4} is classified according to face detectors based on different general detectors, such as fast RCNN, SSD, Yolo, etc. The data in the table is obtained from the official website of WiderFace.
\par
And the precision-recall (PR) curves of our YOLO-FaceV2 face detector, along with the competitors, are shown in Figure \ref{Fig.6}.

\section{Conclusion}
In this paper, aimed to address the problems of varying face scales, easy and hard sample imbalance and face occlusion, we propose a YOLOv5-based face detection method called YOLO-FaceV2. For the problems of varying face scales, we fuse the P2 layer into the feature pyramid to improve the resolution of small objects, design the RFE module to enhance the receptive field and use NWD Loss to improve the robustness of our model to small target detection. And we introduce Slide function to alleviate the easy and hard sample imbalance. For the face occlusion, we use SEAM module and Repulsion Loss to solve it. Beside, We use the information of the effective receptive field to design the anchor. Finally, we achieve close to or exceeding SOTA performance on the WiderFace validation Easy and Medium subsets.

\bibliographystyle{unsrt}
\bibliography{ref}

\end{document}